\documentclass{article}

\usepackage[final,dblblindworkshop]{neurips_2026}
\workshoptitle{SocialAgent: Second Workshop on Large Language Models for Social Reasoning and Simulation}
\usepackage[utf8]{inputenc}
\usepackage[T1]{fontenc}
\usepackage{booktabs}
\usepackage{amsmath}
\usepackage{microtype}
\usepackage{hyperref}
\usepackage{url}
\usepackage{xcolor}
\usepackage{fvextra}

\title{Structuring Survey Histories for LLM Digital Twins:\\A Controlled Study of Background, Decision Procedure, and Evaluation Profiles}

\author{Iris Ye$^\ast$ \quad Tianze Deng$^\ast$ \quad Ozan Candogan$^\dagger$\\
{\footnotesize $^\ast$Equal contribution. \quad $^\dagger$Faculty advisor.}}

\newcommand{\bde}{\textsc{BDE}}
\newif\ifshowtodo\showtodofalse 

\begin{document}
\maketitle
{\makeatletter\renewcommand{\@makefntext}[1]{\noindent#1}\makeatother\begin{NoHyper}\footnotetext{Booth School of Business, University of Chicago.\\Contact: \texttt{\{iris.ye,tianze10,Ozan.Candogan\}@chicagobooth.edu}.}\end{NoHyper}}

\begin{abstract}
LLM-based digital twins aim to predict how individuals would respond in new settings from representations of their prior responses. Existing evidence that compression of survey histories need not reduce predictive accuracy suggests that representation structure, rather than information volume alone, may be an important design choice. We study \bde{}, a fixed representation that organizes each respondent's survey history into Background, Decision procedure, and Evaluation. We evaluate it on Twin-2K-500~\citep{toubia2025twin2k500} by predicting each respondent's held-out answers on 17 tasks. In a paired evaluation with 100 respondents, a fully inferred \bde{} profile improves overall accuracy over the raw survey transcript by 1.27 percentage points (pp) on \texttt{gpt-5.4-nano} ($p<0.01$) and 1.19 pp on \texttt{gpt-5.4-mini} ($p<0.001$). Three analyses then examine the schema, extraction depth, and component composition. Against a length-matched free-form profile written from the same transcript by the same extractor, \bde{} improves pricing accuracy on both simulators (2.15 pp on \texttt{nano}, $p<0.01$; 1.38 pp on \texttt{mini}, $p<0.05$), while its overall advantage is significant only on \texttt{nano} (1.05 pp, $p<0.05$). Allowing deeper inference during extraction yields no statistically significant gain on either simulator. By contrast, adding Evaluation to a Background--Decision procedure profile improves overall accuracy by 1.44 pp on \texttt{nano} ($p<0.001$) and 1.22 pp on \texttt{mini} ($p<0.001$). These results indicate that the advantage over raw transcripts is not explained by compression or length alone, and attribute it primarily to profile composition rather than deeper extraction.

\end{abstract}

\section{Introduction}

LLM-based digital twins predict an individual's answers to new questions from prior survey responses. Twin-2K-500 provides a particularly information-rich setting, in which each of 2,058 respondents answers 500 questions across three waves, of which 88 questions from 17 tasks are held out for evaluation~\citep{toubia2025twin2k500}. Evidence on compressing such histories is mixed: summarizing interview transcripts barely reduces predictive accuracy~\citep{park2024interview}, whereas a Twin-2K-500 persona summary trails the full transcript~\citep{toubia2025twin2k500}. This suggests that how respondent information is organized, not only how much of it is retained, shapes predictive accuracy.

We study \bde{}, a fixed schema that partitions a respondent's survey evidence into three profiles: Background (\texttt{bg}; identity and knowledge), Decision procedure (\texttt{dp}; how the respondent approaches choices), and Evaluation (\texttt{ep}; preferences and priorities). An otherwise fixed simulator uses these profiles to predict held-out responses. We ask two questions. First, does \bde{} improve prediction relative to the raw transcript? Second, if it does, which design choice accounts for the gain: compressing the transcript, imposing the three-part partition, the depth of inference permitted during extraction, or the profiles the representation includes?

The evidence comes from a controlled analysis of 100 respondents with two simulators, \texttt{gpt-5.4-nano} and \texttt{gpt-5.4-mini} (\texttt{nano} and \texttt{mini} below). We compare \bde{} with the raw transcript and with a free-form profile of the same length, and, within \bde{}, vary the extraction depth and which of the three profiles are included. The fully inferred \bde{} representation is modestly more accurate than the raw transcript. It also remains more accurate than the free-form profile overall on \texttt{nano} and in pricing on both simulators, so its benefit is not explained by compression or length alone. Most of this gain comes from \texttt{ep}, and deeper inference during extraction yields no detectable improvement.

\section{BDE Persona Representation}

Each \bde{} profile is extracted from the respondent's survey transcript, which never includes the held-out answers, and is fixed before simulation. 

\paragraph{Background (\texttt{bg}).}
This profile records who the respondent is, what the respondent knows, and what the respondent believes. Inputs include demographics, Big Five traits, values, political and policy beliefs, tested financial and probabilistic knowledge, and indicators of response reliability.

\paragraph{Decision procedure (\texttt{dp}).}
This profile records how the respondent reasons under uncertainty. Inputs include need for cognition, cognitive-test error patterns, need for closure, maximizing tendencies, intertemporal-choice patterns used as reasoning evidence, and calibrated statements about likely susceptibility to specific biases.

\paragraph{Evaluation (\texttt{ep}).}
This profile records what the respondent values. Inputs include budget constraints, consumer orientation, risk and time preferences, social preferences, and input-side willingness-to-pay responses.

The partition is motivated by distinctions in consumer and decision research among identity and beliefs, reasons and decision strategies, and the construction of preferences~\citep{ajzen1991theory,westaby2005behavioral,payne1993adaptive,bettman1998constructive,falk2018global}. These connections motivate the schema rather than establish the psychological validity of any recovered profile, and the present experiments test its predictive consequences.

Extraction depth is varied at three levels. V1 \bde{} permits direct extraction and reallocation of stated evidence. V2 \bde{} additionally permits explicitly labeled demographic and cross-scale inference. V3 \bde{} further allows broader inference and an explicit assessment of self-report reliability. At each depth, \bde{} uses markdown headers for its three sections.

Two further conditions serve as controls. V3 free-form is generated by the same extractor from the same transcript under the same V3 inference rules and is matched to V3 \bde{} in length for each respondent, but it lacks the \bde{} partition. V3 unlabeled consists of the V3 \bde{} text with its section headers removed and all other text unchanged. Appendix~\ref{app:prompts} specifies the constructions.

\section{Experimental Design}
\paragraph{Experimental setup.}
All experiments use Twin-2K-500~\citep{toubia2025twin2k500}. Because the between-subject experiments assign one version to each respondent, each respondent's held-out block contains 63 of the benchmark's 88 held-out questions, from all 17 tasks. These questions yield 120 scored items, 80 from the cognitive-bias experiments and 40 from the pricing study. We report three metrics, computed with the same scoring procedure for every condition. \emph{Overall} averages accuracy over all 120 items, \emph{Cognitive bias} averages over the 80 cognitive-bias items, and \emph{Pricing} over the 40 pricing items. Condition differences are reported in percentage points (pp), and every paired contrast uses the same respondents and held-out questions.

We use \texttt{gpt-5.4-nano} and \texttt{gpt-5.4-mini} as simulators. Both use high reasoning effort with default sampling, and each prompt ends with the respondent's own question block which instructs the model to answer as the described persona in a fixed format (Appendix~\ref{app:simulation}). Across conditions, only the persona representation changes. The evaluation uses $n=100$ respondents.

\paragraph{Contrasts.}
The end-to-end contrast compares V3 \bde{} with the raw transcript and jointly manipulates compression, schema, extraction depth, and composition. The free-form comparison compares V3 \bde{} with V3 free-form and manipulates the schema while holding compression, extraction depth, and length constant. The depth ablation compares V2 \bde{} and V3 \bde{} with V1 \bde{} and manipulates extraction depth while holding schema and composition constant. The component ablation compares subsets of the three profiles at V3 depth and manipulates composition while holding schema and extraction depth constant.

\paragraph{Inference.}
Each contrast $c_B - c_A$ is the mean over respondents of the paired difference in accuracy between a comparison condition $c_B$ and a baseline $c_A$. We report 95\% percentile intervals and two-sided $p$-values from a paired bootstrap over respondents with 10,000 resamples.

\section{Results}

\subsection{End-to-end effect}

V3 \bde{} outperforms the raw transcript by 1.27 pp overall on \texttt{nano} ($p<0.01$) and 1.19 pp on \texttt{mini} ($p<0.001$), with significant pricing gains on both simulators and a significant cognitive-bias gain on \texttt{nano} (Table~\ref{tab:endtoend}).

\begin{table}[t]
\caption{End-to-end effect at $n=100$. V3 \bde{} is compared with the raw survey transcript. Entries are accuracy differences in percentage points with 95\% percentile bootstrap intervals. Stars reproduce the source paired-bootstrap tests, with $^{*}p<0.05$, $^{**}p<0.01$, and $^{***}p<0.001$, and ns indicates that the interval includes zero.}
\label{tab:endtoend}
\centering
\footnotesize
\setlength{\tabcolsep}{3.2pt}
\begin{tabular}{llccc}
\toprule
Model & Contrast & Overall & Cog. bias & Pricing \\
\midrule
\texttt{nano} & V3 \bde{} $-$ raw & $+1.27^{**}$\,[0.41,\,2.14] & $+0.84^{**}$\,[0.31,\,1.37] & $+1.85^{*}$\,[0.08,\,3.65] \\
\texttt{mini} & V3 \bde{} $-$ raw & $+1.19^{***}$\,[0.50,\,1.89] & $+0.53$ ns\,[$-$0.00,\,1.02] & $+2.10^{**}$\,[0.53,\,3.67] \\
\bottomrule
\end{tabular}
\end{table}

\subsection{Free-form comparison}

The free-form contrast compares V3 \bde{} with V3 free-form, a profile without the three-part partition, holding the extractor, inference rules, and length fixed (Table~\ref{tab:freeform}). Lengths are matched for each respondent, with a median word ratio of 1.00 and a range of 0.90 to 1.10. V3 \bde{} is more accurate overall by 1.05 pp on \texttt{nano} ($p<0.05$) and by 0.57 pp on \texttt{mini}, where the interval includes zero. The difference is concentrated in pricing, where V3 \bde{} is more accurate by 2.15 pp on \texttt{nano} ($p<0.01$) and 1.38 pp on \texttt{mini} ($p<0.05$), while the cognitive-bias difference is small and not statistically significant on either simulator, with both intervals within 0.75 pp of zero. Deleting the section headers from the V3 \bde{} text, by contrast, produces no statistically significant difference on any metric (Appendix~\ref{app:labeling}).

\subsection{Depth ablation}

The depth ablation compares V2 \bde{} or V3 \bde{} with V1 \bde{}, holding composition fixed (Table~\ref{tab:depth}). No contrast is statistically significant on either simulator. The overall differences range from $+0.06$ to $+0.83$ pp on \texttt{nano} and from $+0.11$ to $+0.27$ pp on \texttt{mini}, and all confidence intervals include zero. Extraction depth therefore does not account for the end-to-end gain.

\begin{table}[t]
\caption{Free-form comparison at $n=100$. V3 \bde{} is compared with V3 free-form, a profile of the same respondent written from the raw transcript by the same extractor following the same V3 inference rules, without the \texttt{bg}--\texttt{dp}--\texttt{ep} partition, and matched to V3 \bde{} in length (Appendix~\ref{app:prompts}). Units and significance markers as in Table~\ref{tab:endtoend}.}
\label{tab:freeform}
\centering
\footnotesize
\setlength{\tabcolsep}{3.2pt}
\begin{tabular}{llccc}
\toprule
Model & Contrast & Overall & Cog. bias & Pricing \\
\midrule
\texttt{nano} & V3 \bde{} $-$ V3 free-form & $+1.05^{*}$\,[0.25,\,1.88] & $+0.21$ ns\,[$-$0.32,\,0.74] & $+2.15^{**}$\,[0.52,\,3.82] \\
\texttt{mini} & V3 \bde{} $-$ V3 free-form & $+0.57$ ns\,[$-$0.05,\,1.21] & $-0.04$ ns\,[$-$0.51,\,0.43] & $+1.38^{*}$\,[0.02,\,2.75] \\
\bottomrule
\end{tabular}
\end{table}

\begin{table}[t]
\caption{Depth ablation at $n=100$. V2 \bde{} or V3 \bde{} is compared with V1 \bde{}, holding profile form and composition fixed. Units and significance markers as in Table~\ref{tab:endtoend}.}
\label{tab:depth}
\centering
\footnotesize
\setlength{\tabcolsep}{3.2pt}
\begin{tabular}{llccc}
\toprule
Model & Contrast & Overall & Cog. bias & Pricing \\
\midrule
\texttt{nano} & V2 \bde{} $-$ V1 \bde{} & $+0.83$ ns\,[$-$0.17,\,1.87] & $+0.63$ ns\,[$-$0.62,\,1.92] & $+1.05$ ns\,[$-$0.50,\,2.67] \\
      & V3 \bde{} $-$ V1 \bde{} & $+0.06$ ns\,[$-$1.13,\,1.24] & $-0.45$ ns\,[$-$2.09,\,1.13] & $+0.72$ ns\,[$-$0.93,\,2.45] \\
\midrule
\texttt{mini} & V2 \bde{} $-$ V1 \bde{} & $+0.27$ ns\,[$-$0.33,\,0.88] & $+0.33$ ns\,[$-$0.27,\,0.92] & $+0.20$ ns\,[$-$1.05,\,1.45] \\
      & V3 \bde{} $-$ V1 \bde{} & $+0.11$ ns\,[$-$0.70,\,0.93] & $-0.17$ ns\,[$-$0.85,\,0.48] & $+0.47$ ns\,[$-$1.07,\,2.02] \\
\bottomrule
\end{tabular}
\end{table}

\subsection{Component ablation}

The component ablation compares profiles that pair \texttt{bg} with \texttt{dp}, \texttt{ep}, both, or neither, holding extraction depth fixed at V3 (Table~\ref{tab:component}). Adding \texttt{ep} to \texttt{bg} improves overall and pricing accuracy on both models, by 1.19 and 2.42 pp on \texttt{nano} and by 1.46 and 3.20 pp on \texttt{mini}, respectively (all $p<0.01$). Adding \texttt{dp} to \texttt{bg} produces no statistically significant overall or pricing gain, although it improves cognitive-bias accuracy by 0.59 pp on \texttt{nano} ($p<0.05$). The same asymmetry holds from the other starting point, since adding \texttt{ep} to \texttt{bg+dp} significantly improves overall and pricing accuracy on both models (all $p<0.001$), whereas adding \texttt{dp} to \texttt{bg+ep} produces no statistically significant gain on any metric.

\begin{table}[t]
\caption{Component ablation at $n=100$. Each row adds one or two profiles to the condition named second, at V3 extraction throughout. Units and significance markers as in Table~\ref{tab:endtoend}.}
\label{tab:component}
\centering
\footnotesize
\setlength{\tabcolsep}{3.2pt}
\begin{tabular}{llccc}
\toprule
Model & Contrast & Overall & Cog. bias & Pricing \\
\midrule
\texttt{nano} & \texttt{bg+dp} $-$ \texttt{bg} & $+0.34$ ns\,[$-$0.27,\,0.94] & $+0.59^{*}$\,[0.13,\,1.05] & $+0.03$ ns\,[$-$1.27,\,1.35] \\
      & \texttt{bg+ep} $-$ \texttt{bg} & $+1.19^{**}$\,[0.42,\,1.96] & $+0.26$ ns\,[$-$0.24,\,0.75] & $+2.42^{**}$\,[0.82,\,4.07] \\
      & \texttt{bg+dp+ep} $-$ \texttt{bg} & $+1.78^{***}$\,[1.05,\,2.52] & $+0.53^{*}$\,[0.03,\,1.03] & $+3.42^{***}$\,[1.95,\,4.92] \\
      & \texttt{bg+dp+ep} $-$ \texttt{bg+dp} & $+1.44^{***}$\,[0.75,\,2.15] & $-0.06$ ns\,[$-$0.55,\,0.43] & $+3.40^{***}$\,[1.90,\,4.95] \\
      & \texttt{bg+dp+ep} $-$ \texttt{bg+ep} & $+0.59$ ns\,[$-$0.03,\,1.21] & $+0.27$ ns\,[$-$0.26,\,0.79] & $+1.00$ ns\,[$-$0.27,\,2.25] \\
\midrule
\texttt{mini} & \texttt{bg+dp} $-$ \texttt{bg} & $+0.08$ ns\,[$-$0.52,\,0.66] & $-0.05$ ns\,[$-$0.58,\,0.45] & $+0.22$ ns\,[$-$0.97,\,1.40] \\
      & \texttt{bg+ep} $-$ \texttt{bg} & $+1.46^{***}$\,[0.75,\,2.19] & $+0.10$ ns\,[$-$0.31,\,0.51] & $+3.20^{***}$\,[1.60,\,4.85] \\
      & \texttt{bg+dp+ep} $-$ \texttt{bg} & $+1.29^{***}$\,[0.63,\,1.98] & $+0.11$ ns\,[$-$0.37,\,0.59] & $+2.82^{***}$\,[1.37,\,4.32] \\
      & \texttt{bg+dp+ep} $-$ \texttt{bg+dp} & $+1.22^{***}$\,[0.56,\,1.91] & $+0.16$ ns\,[$-$0.39,\,0.71] & $+2.60^{***}$\,[1.27,\,4.00] \\
      & \texttt{bg+dp+ep} $-$ \texttt{bg+ep} & $-0.17$ ns\,[$-$0.78,\,0.43] & $+0.01$ ns\,[$-$0.44,\,0.44] & $-0.38$ ns\,[$-$1.52,\,0.77] \\
\bottomrule
\end{tabular}
\end{table}

\section{Discussion and Conclusion}
Relative to raw survey transcripts, \bde{} improves accuracy on both simulators, and each ablation targets one candidate source of this gain. The free-form comparison shows that the schema contributes beyond compression and length, since a length-matched profile without the \texttt{bg}--\texttt{dp}--\texttt{ep} partition is less accurate overall on \texttt{nano} and in pricing on both simulators, whereas deleting only the section headers has no detectable effect. The depth ablation detects no effect of extraction depth. The component ablation locates the gain in \texttt{ep}, whose addition to \texttt{bg+dp} improves overall and pricing accuracy on both simulators ($p<0.001$), whereas \texttt{dp} adds no significant overall gain. The benefit of \bde{} is therefore associated with the evaluative content its schema extracts, although the results do not establish that its three-part decomposition is uniquely optimal or that the recovered profiles are psychologically causal.

Four limitations bound these conclusions. First, the evaluation uses one respondent panel, one benchmark, and two models from one family. Second, the free-form comparison changes what is extracted together with how it is organized, so it does not separate the two. Third, the component ablation adds evaluative evidence without an equal-length control, so it does not separate \texttt{ep}-specific content from additional relevant information. Fourth, the Overall score weights pricing as a third of all items rather than as 1 of the benchmark's 17 tasks. Under task averaging, only the end-to-end Overall advantage on \texttt{mini} remains significant, so the Overall gains rest largely on the pricing results, which are unchanged (Appendix~\ref{app:taskavg}).

\bibliographystyle{plainnat}
\bibliography{references}

\appendix

\section{Extraction Depths and Prompts}
\label{app:prompts}

The experiments use three extraction depths, summarized in Table~\ref{tab:promptspec}. Each depth has its own set of extraction prompts, which set how far the extractor may go beyond the evidence the respondent stated.

\begin{table}[h]
\caption{The three extraction depths. Depth sets how far the extractor may infer beyond the respondent's stated answers.}
\label{tab:promptspec}
\centering
\small
\begin{tabular}{lp{0.7\linewidth}}
\toprule
Depth & Allowed inference \\
\midrule
V1 & Stated evidence and direct reallocation of that evidence \\
V2 & V1 plus explicitly labeled demographic and cross-scale inference \\
V3 & V2 plus broader inference and an explicit assessment of self-report reliability \\
\bottomrule
\end{tabular}
\end{table}

We reproduce the V3 \texttt{dp} prompt in full. The V1 and V2 prompts follow the same structure with fewer sections and differ from it mainly in their inference rules, which we reproduce below. The \texttt{bg} and \texttt{ep} prompts use the same inference rules at each depth but cover different constructs and request different sections. Each profile is extracted in its own call to \texttt{gpt-4.1-mini} at temperature 0. The call receives the full input-side transcript in place of \verb|{transcript}| and never the held-out answers.

\paragraph{Decision procedure, V1 (direct), inference rules.}
{\footnotesize
\begin{verbatim}
STRICT RULE: Use ONLY questions that directly and explicitly measure the
reasoning constructs below. Do NOT infer from demographics. Do NOT guess
from related answers. If no direct evidence exists for a component, say so
explicitly rather than fabricating an answer.
\end{verbatim}
}

\paragraph{Decision procedure, V2 (inferred), inference rules.}
{\footnotesize
\begin{verbatim}
RULES:
- Direct evidence: cite the scale name and item number
- Demographic inference: allowed but label it explicitly as:
  "Inferred from [demographic fact]: [conclusion] because [reasoning]"
- Cross-scale inference: allowed when two or more scales point in the same
  direction — note which scales agree
- No speculation beyond what the data can reasonably support
\end{verbatim}
}

\paragraph{Decision procedure, V3 (maximum).}
{\footnotesize
\begin{verbatim}
You are extracting the richest possible reasoning profile from a digital
twin survey transcript. Use every available signal — direct scales,
cognitive test performance and specific error patterns, behavioral game
choices, free-response self-descriptions, somatic symptoms, social
desirability scores, word associations if present, and full demographic
context. Infer aggressively but always state your reasoning chain.

This file describes how the person reasons under uncertainty. It is
descriptive only — do NOT output any scalar prediction, social-influence
weight, or committed numeric forecast at the end. Any influenceability or
bias prediction should be expressed qualitatively in the prose.

Goal: produce the most complete, behaviorally predictive reasoning profile
for simulating this person on judgment-under-uncertainty tasks (framing,
anchoring, base-rate, conjunction, sunk cost, probability matching, myside
bias) and on numeric prediction tasks.

CRITICAL EXTRACTION PRINCIPLES:

1. STATED VS REVEALED GAP — surface any conflict between stated attitudes
   and behavioral choices. Do NOT average. The gap is the signal.

2. COGNITIVE ERROR PATTERNS MATTER MORE THAN TOTAL SCORES — name specific
   error types when they occur.

3. SOMATIC-VS-SELF-REPORT CALIBRATION — if anxiety symptoms are high but
   self-reported Neuroticism is low, note this as a self-report reliability
   issue.

4. FREE-RESPONSE TEXT IS HIGH-SIGNAL — quote or paraphrase when they
   illuminate reasoning style.

5. ACKNOWLEDGE PREDICTIVE LIMITS — per Stanovich & West (2008), most H&B
   biases are NOT reliably correlated with cognitive ability.

The survey covers: Big Five (44 items), Need for Cognition (18 items),
Need for Closure (15 items), Maximizer (6 items), Self-monitoring (13
items), Self-concept clarity (12 items), Values (24 items), Minimalism
(12 items), Empathy (20 items), consumerism/uniqueness scales, economic
games (dictator, ultimatum, trust), intertemporal choice tasks,
lottery/risk tasks, cognitive tests, word association chain, social
desirability, and demographics.

Output EXACTLY this format with no other text:

---DECISION_PROCEDURE---
Write prose in these sections, using markdown ## headers. No bullet lists in
the body prose.

## Effort regulation and cognitive style
Full synthesis of NFC score, cognitive test performance with named error
patterns, Conscientiousness from Big Five, and any free-response evidence
of thinking style. Name specific items and answers. Infer analytical depth
with reasoning.

## Closure and uncertainty tolerance
NFCC scale plus behavioral signals plus free-response evidence of how the
person handles ambiguity in daily life.

## Optimization strategy
Maximizer plus domain-specific evidence. Include evidence from consumer
choices, free-response text about indecision or completion, and any
observable pattern of re-evaluation vs. commitment in the survey itself.

## Self-concept stability
SCC scale, contradictions between self-descriptions, gaps between
aspirational-self and actual-self free responses, any Big Five items
contradicting each other.

## Behavioral vs. stated-preference gap
Thorough analysis. Compare economic games to stated values. Compare free-
response self-descriptions to personality items. Use social desirability
score to weight self-reports globally — if social desirability is high,
all self-reports should be discounted and behavioral data weighted more
heavily.

## Intertemporal patience
Full analysis from delayed-reward items, Conscientiousness, and free-
response evidence of planning vs. procrastination.

## Egocentric projection tendency
Predicts false-consensus magnitude. Use: self-concept clarity (low SCC →
more projection), empathy (low empathy → more projection), political
identity strength (stronger identity → more projection on policy items).

## Bias susceptibility predictions
For each bias, give a prediction and its confidence:

HIGH CONFIDENCE (predictable from cognitive signals):
- Probability matching vs. maximizing
- Dominator (denominator) neglect
- Belief bias
- Wason matching bias

LOW CONFIDENCE (not reliably predictable — acknowledge limits):
- Framing effects
- Anchoring
- Base-rate neglect
- Conjunction fallacy
- Outcome bias
- Sunk-cost effect
- Less-is-more effect
- Omission bias
- Allais / certainty effects
- Myside bias

MODERATE CONFIDENCE (predictable from belief content + projection
tendency):
- False consensus effect
- Risk-benefit nonseparability (affect heuristic)

## Self-report reliability weighting
Explicitly state how much the reader should trust this person's self-
reports given: social desirability score, stated-vs-revealed gaps
observed, somatic-vs-self-report calibration.

## Summary
Four to six sentences capturing the complete reasoning profile. Include
the most predictive single feature (often the stated-vs-revealed gap or a
specific cognitive error pattern).

TRANSCRIPT:
{transcript}
\end{verbatim}
}

\paragraph{Output forms.}
The \bde{} profile concatenates the three profiles in the order \texttt{bg}, \texttt{dp}, \texttt{ep} under the headers \verb|### Background|, \verb|### Decision Procedure|, and \verb|### Evaluation Profile|. V3 unlabeled is derived from it deterministically, without a model call, by deleting every markdown header line except the container line \verb|## Persona Skill Profile:|, so the two forms share their text and differ in length only by the deleted headers.

\paragraph{V3 free-form.}
V3 free-form is written by the same extractor at temperature 0 in three calls that follow the order of the survey. Each call sees the full transcript but describes only its third, and the parts are joined without headers. Each part targets an equal share of the respondent's V3 \bde{} word count, and a part outside its window is revised in the same conversation, up to four times: a short part receives a follow-up asking for the missing words as new paragraphs, which are appended, and a long part is rewritten at its target length. Sentences that summarize or refer to parts are deleted before each length check. The prompt for each part is below.

{\footnotesize
\begin{verbatim}
You are writing the richest possible profile of a person from a digital
twin survey transcript. Use every available signal — direct scales, free-
response text, behavioral patterns, demographic context, and any cross-scale
convergence. Infer aggressively but always state your reasoning chain.

Goal: produce the most complete description of this person, for use by a
downstream simulation that will answer new questions as if it were this
person.

CRITICAL EXTRACTION PRINCIPLES:

1. ACQUIESCENCE IN VALUES RATINGS — name it when present; extract only
   relative structure.

2. FREE-RESPONSE TEXT IS HIGH-SIGNAL — aspirational and actual-self
   descriptions often reveal self-concept that scales miss. Quote or
   paraphrase them when informative.

3. SELF-REPORT CONTRADICTIONS ARE DATA — if self-reported Neuroticism is
   low but somatic anxiety symptoms are moderate-to-severe, this is a
   finding about self-awareness, not a contradiction to smooth over.

4. DOMAIN FAMILIARITY MUST BE INFERRED RESPONSIBLY — use education,
   income, employment, household, and region to condition what topics the
   person is likely to know about. State confidence for each inference.
   Do not fabricate a biography.

The survey covers: Big Five (44 items), Need for Cognition (18 items),
Need for Closure (15 items), Maximizer (6 items), Self-monitoring (13
items), Self-concept clarity (12 items), Values (24 items), Minimalism
(12 items), Empathy (20 items), consumerism/uniqueness scales, economic
games, cognitive tests, word association chain, social desirability,
full demographics, and free-response items.

The profile is written in 3 consecutive parts that follow the order of
the survey. The transcript below is marked into 3 parts with
<<<PART k START>>> and <<<PART k END>>> lines. Write PART {k} of the profile:
describe what the questions inside PART {k} reveal about this person. You may
draw on the rest of the transcript as context and for cross-scale inference,
but do not describe material whose evidence lies in the other parts; those
parts are written separately. The parts will be joined into one continuous
profile, so the reader must not be able to tell where one part ends:
- Do not write an introduction, a summary, or a conclusion.
- Never refer to parts, sections, or the survey's structure (no "this
  section", "Part 1", "earlier in the survey").
- Refer to the person as {pronoun} throughout.

Output EXACTLY this format with no other text:

---SUMMARY---
Plain prose paragraphs. Decide for yourself what to include and how to
organize it. Do not use markdown headers, section titles, labels, part
numbers, or bullet lists.

LENGTH: about {target_words} words. Stay within {lo_words}-{hi_words} words.

TRANSCRIPT:
{transcript}
\end{verbatim}
}

\section{Simulation Prompt}
\label{app:simulation}

The simulator answers all held-out questions for a respondent in a single call. Each prompt opens with the persona representation, introduced by \verb|## Persona Skill Profile:| or, for the raw transcript, \verb|## Persona Profile (This individual's past survey responses):|, followed by the respondent's masked question block and JSON instructions, which are identical across conditions. Unparseable responses are requested again, up to 10 times. The excerpt below shows one prompt, with omissions in square brackets. The placeholder \verb|[persona representation]| stands for the text of the condition being evaluated, which is the profile itself for every \bde{} variant, V3 free-form, and V3 unlabeled, and the respondent's full input-side transcript for the raw condition.

{\footnotesize
\begin{Verbatim}[breaklines=true,breakanywhere=true]
## Persona Skill Profile:
[persona representation]

---
## New Survey Question & Instructions (Please respond as the persona described above):
Please answer the following questions as if you were taking this survey. The expected output is a JSON object and the format is provided in the end. 
---

Q1:
Would you support or oppose...
Question Type: Matrix
Options:
  1 = Strongly oppose
  [... options 2-5 omitted ...]
1. Placing a tax on carbon emissions?
Answer: [Masked]
[... items 2-10 of Q1 and questions Q2-Q23 omitted ...]

You will now be asked a series of 40 questions. [... introduction to the pricing study omitted ...]

Q24:
Please consider the following product category: Refrigerated Entrees. Suppose you are in a grocery store, and you see the following product in that category: John Soules Foods Chicken Breast Fajita Strips, Refrigerated, 16oz, 18g Protein per 3oz Serving Size. The product is priced at: $11.96. Would you or would you not purchase this product?
Question Type: Single Choice
Options:
  1 - Yes, I would purchase the product
  2 - No, I would not purchase the product
Answer: [Masked]
[... questions Q25-Q63 omitted ...]

### Format Instructions:
In order to facilitate the postprocessing, you should generate string that can be parsed into a valid JSON object with the following format:
{
    "Q1": {
    "Question Type": "XX",
    "Answers": {
        see below
    } 
    },
    ....
}

[... answer formats for Matrix, Single Choice, Slider, and Text Entry questions omitted ...]
\end{Verbatim}
}

\section{Labeling Control}
\label{app:labeling}

Table~\ref{tab:form} compares V3 \bde{} with the V3 unlabeled profile, which is the same text with the header lines deleted and is 3.3 per cent shorter as a result. No metric is statistically significant on either simulator, and the two simulators disagree in sign. The overall differences are $+0.52$ pp on \texttt{nano} and $-0.31$ pp on \texttt{mini}, with paired correlations between 0.79 and 0.93 and intervals no wider than about $\pm 1.2$ pp.

\begin{table}[h]
\caption{Labeling comparison at $n=100$. V3 \bde{} is compared with the V3 unlabeled profile, which is the same text with every section header deleted. Units and significance markers as in Table~\ref{tab:endtoend}.}
\label{tab:form}
\centering
\footnotesize
\setlength{\tabcolsep}{3.2pt}
\begin{tabular}{llccc}
\toprule
Model & Contrast & Overall & Cog. bias & Pricing \\
\midrule
\texttt{nano} & V3 \bde{} $-$ V3 unlabeled & $+0.52$ ns\,[$-$0.07,\,1.12] & $+0.36$ ns\,[$-$0.24,\,0.98] & $+0.75$ ns\,[$-$0.45,\,2.00] \\
\texttt{mini} & V3 \bde{} $-$ V3 unlabeled & $-0.31$ ns\,[$-$0.90,\,0.26] & $-0.20$ ns\,[$-$0.67,\,0.27] & $-0.43$ ns\,[$-$1.57,\,0.72] \\
\bottomrule
\end{tabular}
\end{table}

\section{Task-Averaged Scores}
\label{app:taskavg}

The main text averages accuracy over the 120 scored items, so the 40 pricing items carry a third of the Overall score. Twin-2K-500 instead averages within each of its 17 tasks and then across tasks, giving the pricing study 1/17 of the score~\citep{toubia2025twin2k500}. Table~\ref{tab:taskavg} re-scores the main contrasts under this rule, with anchoring deciles taken from the wave 2 answers of all 2,058 respondents. Pricing is one task under either rule and is unchanged, and the depth ablation is not re-scored. Under task averaging, V3 \bde{} remains more accurate than the raw transcript only on \texttt{mini}, and no Overall advantage over free-form or from adding \texttt{ep} is statistically significant. Adding \texttt{ep} to \texttt{bg+dp} raises pricing accuracy by 3.40 and 2.60 pp but lowers accuracy on two two-item tasks, by 7.35 and 11.40 pp on sunk cost and 4.00 and 3.00 pp on absolute versus relative savings (\texttt{nano} and \texttt{mini}; Table~\ref{tab:pertask}), and these tasks carry the same weight as the pricing study once tasks are averaged.

\begin{table}[h]
\caption{Main contrasts re-scored with the Twin-2K-500 task-averaged score at $n=100$. Overall averages over the 17 tasks and Cog.\ bias over the 16 cognitive-bias tasks; pricing is unchanged from the main text. Units and significance markers as in Table~\ref{tab:endtoend}.}
\label{tab:taskavg}
\centering
\footnotesize
\setlength{\tabcolsep}{3.2pt}
\begin{tabular}{llcc}
\toprule
Model & Contrast & Overall & Cog. bias \\
\midrule
\texttt{nano} & V3 \bde{} $-$ raw & $+0.27$ ns\,[$-$0.75,\,1.27] & $+0.17$ ns\,[$-$0.91,\,1.21] \\
 & V3 \bde{} $-$ V3 free-form & $+0.22$ ns\,[$-$0.85,\,1.24] & $+0.09$ ns\,[$-$1.03,\,1.17] \\
 & \texttt{bg+ep} $-$ \texttt{bg} & $+0.67$ ns\,[$-$0.41,\,1.79] & $+0.56$ ns\,[$-$0.56,\,1.72] \\
 & \texttt{bg+dp+ep} $-$ \texttt{bg+dp} & $-0.12$ ns\,[$-$1.12,\,0.83] & $-0.34$ ns\,[$-$1.40,\,0.67] \\
\midrule
\texttt{mini} & V3 \bde{} $-$ raw & $+1.21^{**}$\,[0.35,\,2.06] & $+1.16^{*}$\,[0.23,\,2.05] \\
 & V3 \bde{} $-$ V3 free-form & $+0.12$ ns\,[$-$0.86,\,1.05] & $+0.04$ ns\,[$-$0.99,\,1.03] \\
 & \texttt{bg+ep} $-$ \texttt{bg} & $-0.38$ ns\,[$-$1.39,\,0.59] & $-0.61$ ns\,[$-$1.67,\,0.43] \\
 & \texttt{bg+dp+ep} $-$ \texttt{bg+dp} & $-0.36$ ns\,[$-$1.39,\,0.64] & $-0.54$ ns\,[$-$1.64,\,0.50] \\
\bottomrule
\end{tabular}
\end{table}

\begin{table}[h]
\caption{Mean per-task accuracy difference from adding \texttt{ep}, \texttt{bg+dp+ep} $-$ \texttt{bg+dp}, in percentage points at $n=100$. Items is the number of scored items in the task.}
\label{tab:pertask}
\centering
\footnotesize
\begin{tabular}{lrrr}
\toprule
Task & Items & \texttt{nano} & \texttt{mini} \\
\midrule
Base rate & 2 & $+0.43$ & $-0.43$ \\
Outcome bias & 2 & $+0.67$ & $-0.33$ \\
Sunk cost & 2 & $-7.35$ & $-11.40$ \\
Allais & 2 & $+4.00$ & $+3.00$ \\
Framing & 2 & $-2.80$ & $-0.60$ \\
Conjunction (Linda) & 6 & $+0.47$ & $+0.40$ \\
Anchoring & 4 & $-0.17$ & $-0.17$ \\
Absolute vs.\ relative savings & 2 & $-4.00$ & $-3.00$ \\
Myside bias & 2 & $-0.40$ & $+1.20$ \\
Less is more & 9 & $+1.28$ & $+0.57$ \\
WTA/WTP (Thaler) & 3 & $+0.44$ & $+0.11$ \\
False consensus & 18 & $+0.24$ & $+0.60$ \\
Nonseparability of risk and benefit & 8 & $-0.65$ & $-0.02$ \\
Omission bias & 1 & $+2.33$ & $+1.67$ \\
Probability matching & 16 & $+0.00$ & $+0.73$ \\
Denominator neglect & 1 & $+0.00$ & $-1.00$ \\
\midrule
Pricing & 40 & $+3.40$ & $+2.60$ \\
\bottomrule
\end{tabular}
\end{table}

\end{document}